\documentclass[conference,a4paper]{IEEEtran}
\IEEEoverridecommandlockouts

\usepackage[hidelinks]{hyperref}
\usepackage[cmex10]{amsmath}
\usepackage{amssymb,amsfonts}
\usepackage{dblfloatfix}
\usepackage{bm}
\usepackage[ruled,vlined]{algorithm2e}
\usepackage{graphicx}
\graphicspath{{Figures/PDF/}{Figures/PNG/}}

\usepackage{booktabs}
\usepackage{siunitx}
\usepackage[numbers,compress]{natbib}
\usepackage{texnames}
\usepackage{bm,bbm}
\usepackage{orcidlink}
\usepackage{tabularray}
\usepackage{titlesec}
\usepackage{colortbl}
\begin{document}

\title{\uppercase{WSSM: Geographic-Enhanced Hierarchical state-space model for Global Station Weather Forecast}}

\author{\textit{Songru Yang}$^1$, \textit{Zili Liu}$^{1,2}$, \\\textit{Zhenwei Shi}$^1$,~\IEEEmembership{Senior Member,~IEEE}, and \textit{Zhengxia Zou}$^{1,*}$,~\IEEEmembership{Senior Member,~IEEE}\\

\\$^1$Beihang University, $^2$Shanghai AI Laboratory

\thanks{
	The work was supported by the National Natural Science
Foundation of China under Grant 62471014, 62125102, and U24B20177, the National Key Research and Development Program of China under Grant 2022ZD0160401, the Beijing Natural Science Foundation under Grant JL23005, and the Fundamental Research Funds for the Central Universities. \emph{(Corresponding author: Zhengxia Zou (e-mail: zhengxiazou@buaa.edu.cn))}}
}
\maketitle
\begin{abstract}
Global Station Weather Forecasting (GSWF), a prominent meteorological research area, is pivotal in providing timely localized weather predictions. Despite the progress existing models have made in the overall accuracy of the GSWF, executing high-precision extreme event prediction still presents a substantial challenge. The recent emergence of state-space models, with their ability to efficiently capture continuous-time
dynamics and latent states, offer potential solutions. However, early investigations indicated that Mamba underperforms in the context of GSWF, suggesting further adaptation and optimization. To tackle this problem, in this paper, we introduce Weather State-space Model  (WSSM), a novel Mamba-based approach tailored for GSWF. Geographical knowledge is integrated in addition to the widely-used positional encoding to represent the absolute special-temporal position. The multi-scale time-frequency features are synthesized from coarse to fine to model the seasonal to extreme weather dynamic. Our method effectively improves the overall prediction accuracy and addresses the challenge of forecasting extreme weather events. The state-of-the-art results obtained on the Weather-5K subset underscore the efficacy of the WSSM. 
\end{abstract}

\begin{IEEEkeywords}
	Meteorology, weather forecasting, state-space model, multivariate time series forecasting
\end{IEEEkeywords}

\section{Introduction}

Global Station Weather Forecasting (GSWF) plays a vital role in providing localized and accurate weather predictions worldwide \cite{wu2023interpretable,han2024weather}. It supports critical sectors such as disaster management, agriculture, and transportation by enabling precise, site-specific forecasts, which are increasingly important in mitigating the impacts of climate change and extreme weather events \cite{stott2016climate}.

Existing data-driven approaches to GSWF typically model the problem as a time series forecasting task. These methods leverage historical weather observations to predict future conditions, often employing advanced time series models such as multilayer perception (MLP) \cite{zeng2023transformers, cyclenet,liu2024deriving}, Long Short-Term Memory (LSTM) \cite{karevan2020transductive} and Transformer \cite{zhou2022fedformer,liu2023itransformer,liu2022pyraformer}. By directly applying these state-of-the-art models, researchers aim to capture the complex temporal dependencies and nonlinear patterns inherent in weather data, offering a promising alternative to traditional physics-based forecasting methods. Despite the effectiveness of existing time series models, these approaches often struggle to accurately model the complex dynamical processes governing atmospheric behavior. Such limitations arise from their limited ability to capture long-range dependencies and underlying physical structures critical for precise weather forecasting \cite{wang2024mamba}.

Recently, state-space models, such as the Mamba model \cite{gu2023mamba, liu2024mambads}, have shown significant potential in time series modeling due to their ability to efficiently capture continuous-time dynamics and latent states. However, early investigations have revealed that Mamba underperforms compared to Transformer-based state-of-the-art methods in the context of GSWF \cite{han2024weather}. This suggests that while Mamba excels in general time series tasks, further adaptation and optimization are required to address the unique challenges posed by the complex dynamical processes inherent in GSWF.

To address the challenges mentioned above, we propose a novel Mamba-based model architecture, namely \emph{WSSM} (Weather State-space Model), tailored specifically to the characteristics of the GSWF task. Specifically, to leverage the acquisition time and location that is highly correlated with the attributes of meteorological signals, we designed a \emph{Geographical encoding} to introduce the geographical information such as acquisition time and location into feature encoding. Meanwhile, meteorological signals exhibit hierarchical characteristics, from macroscopic trends to microscopic fluctuations under different time resolutions. We have designed a shared encoder named \emph{Hierarchical Bi-Mamba encoder}, which comprehensively extracts the coarse to fine features of hierarchical sequences in a multi-resolution bidirectional scanning manner. Finally, we highlight the frequency characteristics of meteorological signals and design a \emph{Time-frequency Bi-Mamba block} as the basic unit of the encoder, to better model the changes from low to high frequency in hierarchical signals through a learnable frequency filtering. By incorporating domain-specific insights and adapting the state-space modeling framework to better capture the intricate temporal and spatial dependencies of weather dynamics, our approach aims to bridge the performance gap with Transformer-based methods while leveraging Mamba's strengths in efficiency and modeling continuous-time processes.

\begin{figure*}[!t]
\centering
\includegraphics[width=\textwidth]{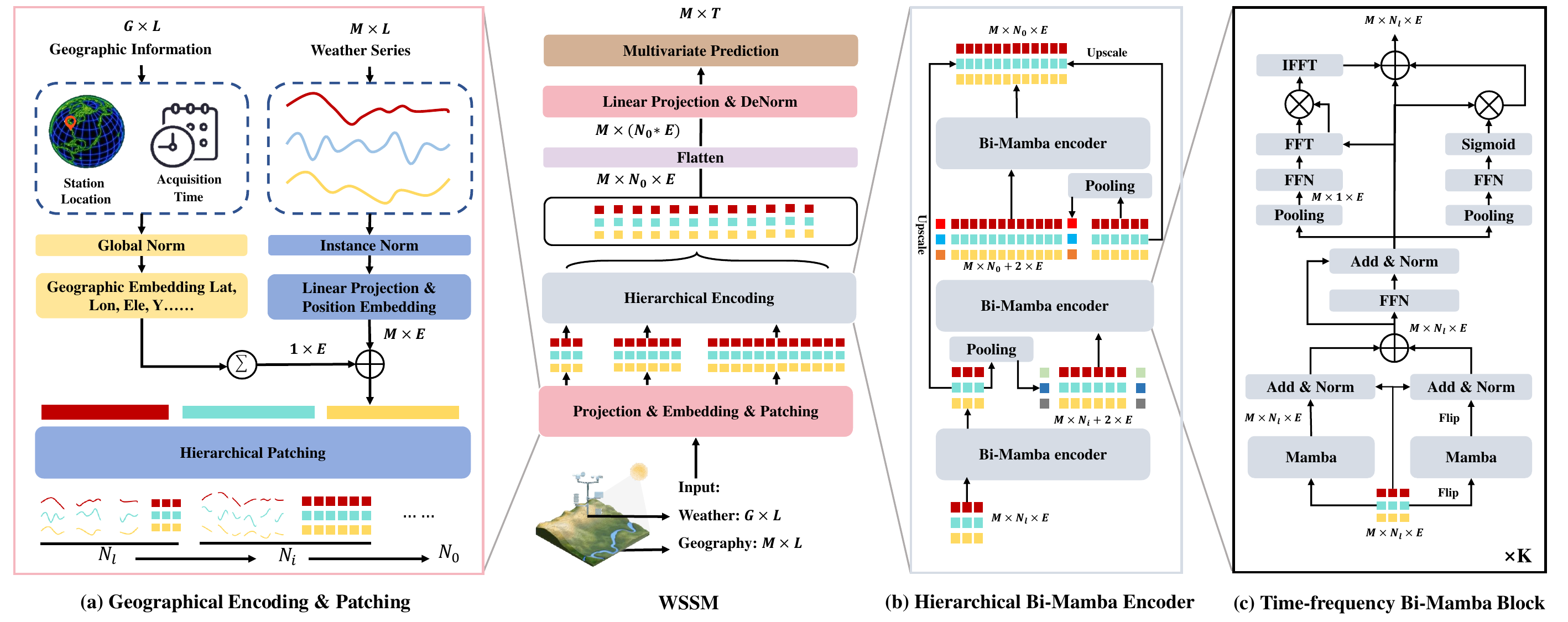}
\caption{The overall framework of WSSM. (a) The Geographical Encoding and Hierarchical Patching process integrate geographical information into weather sequences with different time resolutions. (b) The Hierarchical Bi-Mamba Encoder uses a shared encoder to fuse hierarchical meteorological sequences. (c) The Time-frequency Bi-Mamba Block based on bidirectional Mamba adds a frequency branch and a multivariable branch to simultaneously encode the time-frequency-multivariable features.}
\label{igrass}
\end{figure*}

The primary contributions of this paper can be encapsulated as follows: 1) We propose a novel Mamba-based method WSSM for Global Station Weather Forecasting which integrates geographical information and hierarchical encoding to facilitate weather dynamics modeling and extreme weather event forecasts. 2)  We design a geographical encoding to embed the acquisition time and location. A hierarchical Bi-Mamba encoder built with Time-frequency Bi-Mamba blocks is further proposed to integrate sequence features at multi-scale. 3) We compare the proposed method with various advanced methods on the Weather-5K \cite{han2024weather} subset containing 100 stations. The results demonstrate that our method achieves state-of-the-art performance with particularly outstanding performance in extreme weather forecasts.


\section{Methodology}

\subsection{Overview}
To enhance the performance of the Mamba model on the task of station weather forecasting, we considered the geographical information relevance of meteorological signals and further designed the WSSM model based on the SSM structure from the perspective of multiscale in time and frequency. In the WSSM model, we put forward three innovation modules: a Geographical encoding, a hierarchical Mamba encoder, and a time-frequency Bi-Mamba block. The overall structure of the WSSM is shown in Fig.~\ref{igrass}. 

\subsection{Geographical encoding}

Most existing methods regard meteorological signals as sliced 1-D time series and ignore the real-world temporal-spatial environment in which these signals are generated, such as acquisition time or sensor location. However, such information is highly relevant to weather dynamics and cannot be reflected within a short observation window. To address this, we propose Geographical encoding, which allows the model to capture geographical-related patterns and increase the accuracy of the predictions. 

Geographical encoding includes time encoding and location encoding. As shown in Fig.~\ref{igrass} (a), for station location, we encode the normalized longitude, latitude, and elevation through 3 independent linear layers to obtain location embeddings $\bm{E}_{lat}$, $\bm{E}_{lon}$, and $\bm{E}_{ele}$. For acquisition time, we encode the timestamps of year, month, day, and hour through 4 independent linear layers to obtain time embeddings $\bm{E}_{y}$, $\bm{E}_{m}$, $\bm{E}_{d}$, and $\bm{E}_{h}$. We perform Geographical encoding on the feature input to the encoder and decoder. The mathematical representation of this process is as follows:
\begin{equation}
\label{GE}
\begin{aligned}
    &\text{GeoEnc}(\bm{x}) = \text{Linear}_{lat,y,\ldots}(\bm{lat}_x,\bm{year}_x,\ldots), \\
    &\bm{h} = \text{Embedding}(\bm{x})+\text{PosEnc}(\bm{x})+\text{GeoEnc}(\bm{x}),\\
\end{aligned}
\end{equation}
where $\bm{x}$ is the input weather sequence, and $\bm{h}$ is the sequence feature input to the encoder and decoder. When encoding the features input to the decoder, future timestamps are used.
\subsection{Hierarchical Bi-Mamba encoder}
\label{Hierarchical Bi-Mamba encoder}
We designed a Hierarchical Bi-Mamba encoder to handle features of meteorological signals at multiple time resolutions. As shown in Fig.~\ref{igrass} (b), firstly, we perform hierarchical patch encoding on the input time series to obtain a set of micro-to-macro features, namely [$\bm{S}_{N_0}$, $\bm{S}_{N_1}$, \ldots, $\bm{\bm{S}_{N_l}}]$ where $N_i$ is the number of patches ($N_0>N_1>\ldots>N_l$). Then, we use the Bi-Mamba encoder to process them sequentially from $N_l$ to $N_0$, enabling the network to learn the patterns in meteorological variables from coarse to fine. Between two successive scales, we insert the average pooled previous features into the start and end positions of the current sequence, so that the encoder can explicitly utilize the global prior of the previous scale when processing the current one. Across all scales, we bilinearly sample the low-scale features to the highest scale as the final encoder output. The mathematical process between two scales can be expressed as follows: 
\begin{equation}
\label{HBME}
\begin{aligned}
    &\bm{S}_{N_i} = \text{Patchify}(\bm{h}, \text{patchsize}_i)\in\mathbb{R}^{M \times N_i \times d},\\
    &\bm{z}_{N_i} = \text{Encoder}(\bm{S}_{N_i})\in\mathbb{R}^{M \times N_i \times d}, \\
    &\bm{hp}_{N_i} = \text{Average}(\bm{z}_{N_i})\in\mathbb{R}^{M \times 1 \times d},\\
    &\bm{S}_{N_{i-1}} = \text{Patchify}(\bm{h}, \text{patchsize}_{i-1})\in\mathbb{R}^{M \times N_{i-1} \times d},\\
    &\bm{S}_{N_{i-1}} = \text{Concat}[\bm{hp}_{N_i}, \bm{S}_{N_{i-1}}, \bm{hp}_{N_i}]\in\mathbb{R}^{M \times N_{i-1}+2 \times d},\\  
    &\bm{z}_{N_{i-1}} = \text{Encoder}(\bm{S}_{N_{i-1}})[1:-1]\in\mathbb{R}^{M \times N_{i-1} \times d}, \\
    &\bm{z} = \text{Upscale}(\bm{z}_{N_{i}}) + \text{Upscale}(\bm{z}_{N_{i-1}})\in\mathbb{R}^{M \times N_0 \times d}, \\
\end{aligned}
\end{equation}
Where $\bm{z_{n_i}}$ is the feature at scale $n_i$, $\bm{hd_{n_i}}$ is the average pooled head patch of scale $n_i$, $\bm{z}$ is the final encoded feature.

Through this process, we compress the multi-scale features into a single encoder, which improves the prediction accuracy.

\subsection{Time-frequency Bi-Mamba block}
Frequency information is more suitable for characterizing the periodicity and extremity that occur simultaneously in meteorological signals by low and high frequency components \cite{wang2024timemixer}\cite{yi2024filternet}. We designed a time-frequency Bi-Mamba block with the ability to extract features across the time domain, the frequency domain, and between multiple variables simultaneously as the basic unit of the Hierarchical Bi-Mamba encoder in \ref{Hierarchical Bi-Mamba encoder}. 

As shown in Fig.~\ref{igrass} (c), firstly, we expand the unidirectional Mamba into a bidirectional Mamba $BM(\cdot)$ to enhance the feature extraction ability \cite{liang2024bi}. Then, we transform the fused time-domain features from the two Mamba blocks into the frequency domain. A learnable filter $F(\cdot)$ is added to adaptively screen the important frequency components. After transforming the features back from the frequency domain to the time domain, we employed a simple interaction MLP $I(\cdot)$ for the information interaction among multiple variables to explore the correlations among the variables. The mathematical process in the time-frequency Bi-Mamba block can be expressed as follows: 
\begin{equation}
\label{TFM}
\begin{aligned}
    &\bm{z}_f, \bm{z}_b = \bm{z}, \text{Flip}(\bm{z})\in\mathbb{R}^{M \times N \times d}, \\
    &\bm{z}_{T_f}, \bm{z}_{T_b} = BM(\bm{z}_f,\bm{z}_b), \\
    &\bm{z}_T = \bm{z}_{T_f}+\text{Flip}(\bm{z}_{T_b}),\\
    &\bm{z}_F = \text{iFFT}(\text{FFT}(\bm{z}_T)\cdot\text{FFT}(F(\text{Average}(\bm{z}_T)))), \\
    &\bm{z}_I = \bm{z}_F\cdot I(\text{Average}(\bm{z}_F))\in\mathbb{R}^{M \times N \times d}, \\
\end{aligned}
\end{equation}
Through this process, frequency is intensively incorporated into feature extraction, which enhances the model's ability to predict sudden changes.

\begin{table*}
\centering
\caption{Forecasting results on our WEATHER-5K subset. We report the results at 3 different prediction lengths: 48, 72, and 120, where the input length is 48.}
\label{main result 1}
\resizebox{\textwidth}{!}{
\begin{tblr}{
  width = \textwidth,
  colspec = {Q[140]Q[60]Q[60]Q[69]Q[52]Q[62]Q[54]Q[54]Q[62]Q[88]Q[52]Q[62]Q[52]Q[52]},
  cells = {c},
  cell{1}{1-2} = {r=2}{},
  cell{1}{3,5,7,9,11,13} = {c=2}{},
  cell{3, 6, 9, 12, 15, 18}{1} = {r=3}{},
  vline{2,3,5,7,9,11,13} = {1-7,7-10,10-13,13-16,16-19,19-20}{},
  hline{1,21} = {-}{0.08em},
  hline{2} = {3-14}{},
  hline{3,6,9,12,15,18} = {-}{},
  rowsep = 0.01pt,
}
Baselines     & {Lead\\Time } & Temperature &       & Dewpoint &       & Wind Rate &      & Wind Direc. &          & Sea Level &       & Overal &     \\
              &               & MAE         & MSE   & MAE      & MSE   & MAE       & MSE  & MAE         & MSE      & MAE       & MSE   & MAE    & MSE \\
FEDformer \cite{zhou2022fedformer}     & 48            & 3.02        & \textcolor{blue}{14.52} & 2.82     & 17.88 & 1.59      & 5.74 & 72.40       & 8675.78  & 3.62      & 29.62 &        16.69&1748.71
     \\
              & 72            & 3.18        & 20.55 & 3.43     & 24.01 & 1.68      & 5.88 & 77.07       & 9636.45  & 4.88      & 51.91 &        18.04&1947.76     \\
              & 120           & 3.65        & 26.36 & 3.91     & 30.45 & 1.76      & 6.52 & 80.89       & 10570.91 & 5.86      & 71.15 &        19.21&2141.07     \\
iTransformer\cite{liu2023itransformer}  & 48            & \textcolor{blue}{2.68}        & 15.04 & 2.81     & 17.26 & 1.56      & 5.16 & \textcolor{blue}{71.06}       & 8740.38  & 3.49      & 29.23 &        \textcolor{blue}{16.32}&1761.41     \\
              & 72            & \textcolor{blue}{2.98}        & 18.39 & 3.16     & 21.31 & 1.65      & 5.85 & \textcolor{blue}{73.63}      & 9087.07  & 4.26      & 41.51 &        \textcolor{blue}{17.14}&1834.83     \\
              & 120           & \textcolor{blue}{3.27}        & 21.51 & 3.50     & 25.29 & 1.68      & 5.99 & 76.67       & 9524.65  & 5.04      & 55.26 &        18.03&1926.54     \\
Pyraformer \cite{liu2022pyraformer}    & 48            & 2.77        & 14.96 & \textcolor{blue}{2.71}     & \textcolor{red}{15.74} & \textcolor{blue}{1.55}      & \textcolor{red}{4.97} & 72.13       & \textcolor{red}{8039.62}  & \textcolor{blue}{3.42}      & \textcolor{red}{26.41} &        16.52&\textcolor{red}{1620.34}     \\
              & 72            & 3.01        & 17.59 & \textcolor{red}{3.02}     & \textcolor{red}{18.84} & 1.60      & \textcolor{red}{5.25} & 74.64       & \textcolor{red}{8364.53}  & \textcolor{red}{4.05}      & \textcolor{red}{35.58} &        17.26&\textcolor{red}{1688.36}     \\
              & 120           & 3.30        & \textcolor{blue}{20.86} & \textcolor{blue}{3.36}     & \textcolor{red}{22.59} & \textcolor{blue}{1.64}      & \textcolor{blue}{5.58} & 76.87       & \textcolor{red}{8652.83}  & \textcolor{red}{4.65}      & \textcolor{red}{45.40} &        \textcolor{blue}{17.96}&\textcolor{red}{1749.45}    \\
DLinear \cite{zeng2023transformers}       & 48            & 3.28        & 19.70 & 3.16     & 19.19 & 1.58      & 5.11 & 71.34       & 8666.90  & 4.02      & 34.59 &        16.68&1749.10     \\
              & 72            & 3.58        & 23.28 & 3.52     & 23.15 & \textcolor{blue}{1.64}      & 5.48 & 73.76       & 8948.80  & 4.56      & 43.58 &        17.41&1808.86     \\
              & 120           & 3.93        & 27.71 & 3.92     & 27.86 & 1.68      & 5.79 & \textcolor{blue}{76.20}       & 9240.80  & 5.08      & 52.91 &        18.16&1871.01     \\
Mamba \cite{gu2023mamba}         & 48            & 2.78        & 15.52 & 2.76     & 16.86 & 1.56      & 5.15 & 72.67       & 8622.35  & 3.67      & 31.01 &        16.69&1738.18     \\
              & 72            & 3.02        & \textcolor{blue}{18.04} & 3.07     & \textcolor{blue}{19.88} & 1.60      & \textcolor{blue}{5.34} & 74.87       & 8786.61  & 4.33      & 41.51 &        17.38&1774.28     \\
              & 128           & 3.83        & 21.16 & \textcolor{red}{3.34}     & \textcolor{blue}{22.71} & \textcolor{red}{1.63}      & \textcolor{red}{5.54} & 76.71       & \textcolor{blue}{8891.67}  & 4.96      & \textcolor{blue}{51.88} &        18.09&1798.59     \\
WSSM (Ours) & 48            & \textcolor{red}{2.54}        & \textcolor{red}{13.74} & \textcolor{red}{2.67}     & \textcolor{blue}{16.11} & \textcolor{red}{1.53}      & \textcolor{blue}{5.11} & \textcolor{red}{70.37}     & \textcolor{blue}{8459.89}  & \textcolor{red}{3.40}      & \textcolor{blue}{27.95} &        \textcolor{red}{16.10}&\textcolor{blue}{1704.56}     \\
              & 72            & \textcolor{red}{2.91}        & \textcolor{red}{17.50} & \textcolor{blue}{3.07}     & 20.61 & \textcolor{red}{1.60}      & 5.59 & \textcolor{red}{73.59}
       & \textcolor{blue}{8660.48}  & \textcolor{blue}{4.19}      & \textcolor{blue}{40.64} &        \textcolor{red}{17.07}& \textcolor{blue}{1748.96}    \\
              & 128           & \textcolor{red}{3.17}       & \textcolor{red}{20.68} & 3.46     & 24.81 & 1.68      & 5.98 & \textcolor{red}{76.01}       & 9358.91  & \textcolor{blue}{4.96}      & 54.92 &        \textcolor{red}{17.85}& \textcolor{blue}{1893.06}    
\end{tblr}}
\end{table*}

\begin{table*}
\centering
\caption{Extreme weather events forecasting results@SEDI (
\%).}
\label{main result 2}
\resizebox{\textwidth}{!}{
\begin{tblr}{
  width = \textwidth,
  colspec = {Q[140]Q[60]Q[60]Q[69]Q[52]Q[62]Q[54]Q[54]Q[62]Q[88]Q[52]Q[62]Q[52]Q[52]},
  cells = {c},
  cell{1}{1-2} = {r=2}{},
  cell{1}{3,5,7,9,11,13} = {c=2}{},
  cell{3, 6, 9, 12, 15, 18}{1} = {r=3}{},
  vline{2,3,5,7,9,11,13} = {1-7,7-10,10-13,13-16,16-19,19-20}{},
  hline{1,21} = {-}{0.08em},
  hline{2} = {3-14}{},
  hline{3,6,9,12,15,18} = {-}{},
  rowsep = 0.01pt,
}
Baselines     & {Lead\\Time } & Temperature &       & Dewpoint &       & Wind Rate &       & Wind Direc. &       & Sea Level &       & Overal &      \\
              &               & 99.5th      & 90th  & 99.5th   & 90th  & 99.5th    & 90th  & 99.5th      & 90th  & 99.5th    & 90th  & 99.5th & 90th \\
FEDformer \cite{zhou2022fedformer}     & 48            & \textcolor{red}{29.58}       & \textcolor{red}{64.88} & \textcolor{blue}{23.07}    & \textcolor{blue}{57.54} & \textcolor{blue}{3.51}      & \textcolor{blue}{15.14} & \textcolor{blue}{0.87}        & \textcolor{blue}{5.18}  & \textcolor{red}{40.71}     & \textcolor{red}{65.84} &       \textcolor{blue}{19.54}&\textcolor{blue}{41.72}     \\
              & 72            & 17.85       & 55.66 & 11.85    & 43.72 & \textcolor{blue}{2.88}      & 11.54 & \textcolor{blue}{0.32}        & 3.45 & 15.15     & 40.95 &        9.61&22.91      \\
              & 120           & \textcolor{red}{16.31}       & \textcolor{blue}{51.28} & 8.45     & 17.13 & \textcolor{blue}{2.74}      & 4.23  & \textcolor{blue}{0.21}        & 0.66  & 6.68      & 28.54 &        6.88&20.37      \\
iTransformer\cite{liu2023itransformer}  & 48            & 26.64       & 59.93 & 19.64    & 54.35 & 2.66      & 14.39 & 0.44        & 5.09  & 33.11     & 57.80 &        16.50&38.31      \\
              & 72            & \textcolor{red}{22.26}       & \textcolor{blue}{56.29} & \textcolor{blue}{15.41}    & \textcolor{blue}{49.22} & 1.94      & \textcolor{blue}{12.57} & 0.20        & \textcolor{blue}{3.81}  & \textcolor{blue}
              {21.94}     & \textcolor{blue}{47.13} &        \textcolor{blue}{12.35}&\textcolor{blue}{33.80}      \\
              & 120           & 14.42       & 50.77 & \textcolor{blue}{11.10}    & \textcolor{blue}{43.38} & 1.18      & \textcolor{blue}{10.85} & 0.13        & \textcolor{blue}{2.83}  & \textcolor{blue}{13.40}     & \textcolor{blue}{37.10} &        \textcolor{blue}{8.05}&\textcolor{blue}{28.99}      \\
Pyraformer \cite{liu2022pyraformer}    & 48            & 13.32       & 56.29 & 12.70    & 51.45 & 1.76      & 10.60 & 0.10        & 1.47  & 14.62      & 40.32 &        8.50&32.03      \\
              & 72            & 9.64        & 51.41 & 9.53     & 44.51 & 1.78      & 9.20  & 0.09        & 1.11  & 9.78      & 29.54 &        6.16&27.15      \\
              & 120           & 6.53        & 46.65 & 4.77     & 35.75 & 1.33      & 7.89  & 0.03        & 0.87  & 6.86      & 20.94 &        3.85&22.42      \\
DLinear \cite{zeng2023transformers}       & 48            & 5.22        & 31.82 & 6.84     & 31.39 & 0.72      & 10.31 & 0.46        & 3.80  & 7.15      & 32.75 &        4.08&22.01      \\
              & 72            & 4.08        & 27.34 & 5.11     & 26.09 & 0.51      & 8.68  & 0.31        & 3.02  & 4.80      & 25.77 &        2.96&18.18      \\
              & 120           & 3.04        & 23.07 & 3.63     & 21.01 & 0.33      & 7.27  & 0.18        & 2.36  & 2.85      & 19.29 &        2.01&14.58      \\
Mamba \cite{gu2023mamba}       & 48            & 19.06       & 57.23 & 19.60    & 55.33 & 2.39      & 12.80 & 0.06        & 3.09  & 26.62     & 52.91 &        13.55&36.27      \\
              & 72            & 13.10       & 52.24 & 13.56    & 48.32 & 1.42      & 9.87  & 0.02        & 1.79  & 15.60     & 41.73 &        8.74&30.79      \\
              & 128           & 6.45        & 43.99 & 6.00     & 38.99 & 0.91      & 7.22  & 0.02        & 1.03  & 6.54      & 27.87 &        3.98&23.82      \\
WSSM (Ours) & 48            & \textcolor{blue}{27.81}       & \textcolor{blue}{63.74} & \textcolor{red}{25.47}    & \textcolor{red}{58.54} & \textcolor{red}{8.41}     & \textcolor{red}{20.61} & \textcolor{red}{4.19}        & \textcolor{red}{9.77}  & \textcolor{blue}{36.93}     & \textcolor{blue}{58.88} &        \textcolor{red}{20.56}&\textcolor{red}{42.31}      \\
              & 72            & \textcolor{blue}{19.23}       & \textcolor{red}{58.08} & \textcolor{red}{19.11}    & \textcolor{red}{51.32} & \textcolor{red}{7.82}      & \textcolor{red}{19.25} & \textcolor{red}{3.39}        & \textcolor{red}{8.52}  & \textcolor{red}{25.70}     & \textcolor{red}{49.01} &        \textcolor{red}{15.05}&\textcolor{red}{37.24}      \\
              & 128           & \textcolor{blue}{15.44}       & \textcolor{red}{52.20} & \textcolor{red}{12.87}    & \textcolor{red}{44.60} & \textcolor{red}{6.19}      & \textcolor{red}{17.32} & \textcolor{red}{2.81}        & \textcolor{red}{7.22}  & \textcolor{red}{17.44}     & \textcolor{red}{39.33} &        \textcolor{red}{10.95}&\textcolor{red}{32.13}
\end{tblr}}
\end{table*}

\section{EXPERIMENTS}
\subsection{Experimental Dataset and Settings}
We carried out experiments on the 100 meteorological stations subset of the WEATHER-5K dataset to substantiate the efficacy of our method. The entire WEATHER-5K dataset contains data on temperature, dewpoint temperature, wind speed, wind direction, and sea level pressure from 5,672 meteorological stations around the world, covering a period of 10 years with an hourly interval. We obtained the subset by randomly sampling from it. 

We set the length of the observation to last 48 hours and predicted all 5 variables in 48, 72, and 120 hours. The train/val/test (2014-2021/2022/2023) are divided by time coverage with in total 7003.3k/866.5k/866.5k samples.

\subsection{Evaluation Protocol and Metrics}
To assess the performance, we utilized widely recognized evaluation metrics, including Mean Absolute Error (MAE), Mean Squared Error (MSE) for overall precision, and Symmetric Extremal Dependence Index (SEDI) for precision on abrupt change.

\subsection{Implementation Details}
\textbf{Architecture Details:} WSSM sets the dimension $d=128$, and the block number $K=2$. The hierarchical patching parameters (window, stride) are (16, 8), (8, 4), and (2, 1). The unidirectional SSM implementation refers to \cite{zeng2024c}. 

\textbf{Training Details:} WSSM employs MSEloss function for training. We utilize the AdamW optimizer with a
learning rate of 0.0001 and a cosine annealing scheduler with
a linear warmup to decay the learning rate. Our batch size is
set to 1024, and the maximum iteration is 60000.

\subsection{Comparison with the State-of-the-Art}
We have compared the proposed WSSM with a series of state-of-the-art time series prediction methods, including FEDformer\cite{zhou2022fedformer}, iTransformer\cite{liu2023itransformer}, Pyraformer\cite{liu2022pyraformer}, DLinear\cite{zeng2023transformers}, and Mamba\cite{gu2023mamba}. The comparative results on MAE and MSE are presented in Tab. \ref{main result 1}. As illustrated, the proposed WSSM achieved
the optimal or suboptimal performance on $>77\% $ evaluation metrics, surpassing the contemporary state-of-the-art method Pyraformer ($69\%$). Moreover, our major advantage lies in the extreme weather forecast. In the comparison of SEDI shown in Table \ref{main result 2}, our method significantly outperforms all other methods by up to $90\%(0.21 \xrightarrow{} 2.81)$.  For variables with more extreme conditions, such as wind direction and wind rate, our prediction has greater advantages in extreme situations but is suboptimal in accuracy. Further, methods that achieve high overall accuracy often underperform in extreme weather events, such as Pyraformer. Conversely, the opposite is true for iTransformer. This is caused by only fitting the low-frequency components and high-frequency variations. Our method can maintain state-of-the-art performance in overall prediction while gaining a significant advantage on SEDI, indicating that we can not only learn the low-frequency trends but also capture the high-frequency fluctuations, which aligns with our original intention of incorporating hierarchical encoding and frequency. The comprehensive results underscore the effectiveness of our improvements to the Mamba-based model for GSWF.

\section{CONCLUSION}
In this paper, we improve the state-space model, enabling it to achieve state-of-the-art performance that surpasses Transformer-based methods in the global station weather forecasting task. Specifically, we have proposed a Geographical encoding, which incorporates real-world time and location information, allowing the model to understand the absolute spatio-temporal position of weather sequences. In addition, we have designed a Time-frequency Bi-Mamba block and further constructed a Hierarchical Bi-Mamba encoder to synthesize the time-frequency-multivariable features of multi-temporal resolution sequences through the hierarchical bidirectional scanning. Experimental results on the Weather-5K subset underscore the effectiveness of our proposed modules. The WSSM brings a great leap in high-accuracy extreme weather event prediction as well as achieves the best overall performance. This innovative approach paves the way for more accurate and efficient global station weather forecasting.
\small
\bibliographystyle{IEEEtranN}
\bibliography{references}

\end{document}